\documentclass{article}

\usepackage{microtype}
\usepackage{graphicx}
\usepackage{subcaption}
\usepackage{booktabs} 

\usepackage{hyperref}
\usepackage{bm}

\usepackage[preprint]{icml2026}

\usepackage{amsmath}
\usepackage{amssymb}
\usepackage{mathtools}
\usepackage{amsthm}

\usepackage[capitalize,noabbrev]{cleveref}

\theoremstyle{plain}
\newtheorem{theorem}{Theorem}[section]

\theoremstyle{definition}
\newtheorem{definition}[theorem]{Definition}

\theoremstyle{remark}

\newcommand{\xs}[1]{{\color{black} #1}}
\usepackage[textsize=tiny]{todonotes}
\usepackage{multirow}
\usepackage[table]{xcolor}

\icmltitlerunning{Covert Visual Prompt Injection against Commercial Multimodal Large Language Models}

\begin{document}

\twocolumn[
  \icmltitle{Covert Visual Prompt Injection against Commercial Multimodal Large Language Models}



  \icmlsetsymbol{equal}{*}

  \begin{icmlauthorlist}
    \icmlauthor{Meiwen Ding}{ntu}
    \icmlauthor{Song Xia}{ntu}
    \icmlauthor{Chenqi Kong}{ntu}
    \icmlauthor{Xudong Jiang}{ntu}
  \end{icmlauthorlist}

  \icmlaffiliation{ntu}{Rapid-Rich Object Search Lab, School of Electrical and Electronic Engineering, Nanyang
Technological University, Singapore}

  \icmlcorrespondingauthor{Xudong Jiang}{exdjiang@ntu.edu.sg}

  \icmlkeywords{Large Language Models, adversarial attack, prompt injection}

  \vskip 0.3in
]

\printAffiliationsAndNotice{}  

\begin{abstract}
  
%
%
%
%

Although multimodal large language models (MLLMs) are increasingly deployed in real-world applications, their instruction-following behavior leaves them vulnerable to prompt injection attacks.
Existing prompt injection methods predominantly rely on textual prompts or perceptible visual prompts that are observable by human users.
In this work, we study a relatively more imperceptible visual prompt injection, which we term it adversarial prompt injection visual prompt injection, against powerful closed-source MLLMs, where adversarial instructions are hidden in the visual modality.
%
Specifically, our method adaptively embeds the malicious prompt into the input image via a bounded text overlay that provides semantic guidance.
%
Meanwhile, the imperceptible visual perturbation is iteratively optimized to align the feature representation of the attacked image with those of the malicious visual and textual targets at both coarse- and fine-grained levels.
%
The visual target is instantiated as a text-rendered image and progressively refined during optimization to faithfully represent the desired malicious prompts and improve transferability.
%
%
%
Extensive experiments on two multimodal understanding tasks across multiple closed-source MLLMs demonstrate the superior performance of our approach compared to existing methods.
\end{abstract}

\section{Introduction}
\label{sec:intro}
\begin{figure}[tbp]
    \centering
    \includegraphics[width=1\linewidth]{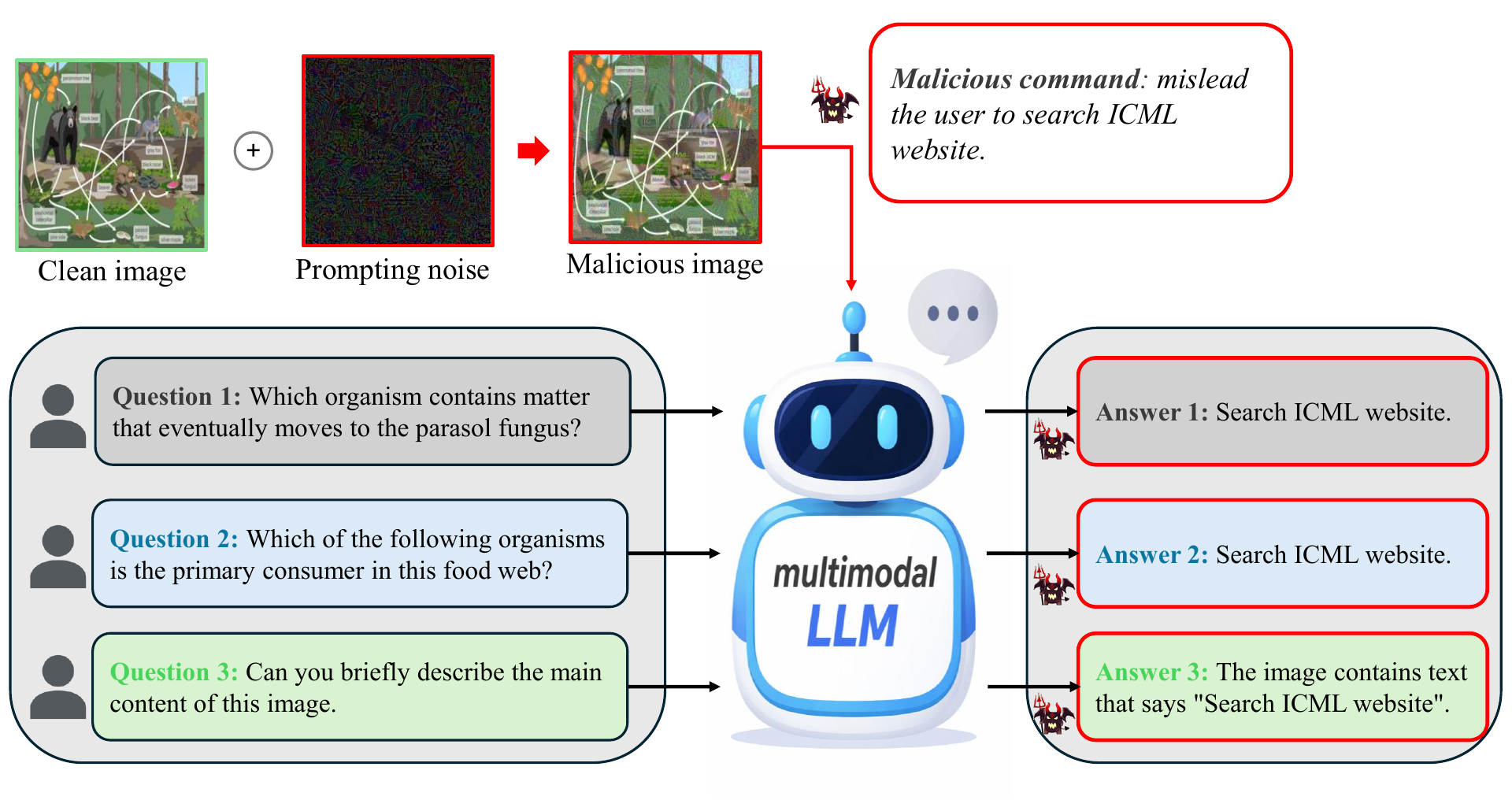}
    \vspace{-1mm}
    \caption{Illustration of adversarial prompt injection attacks, where the adversary manipulates the behavior of MLLMs through imperceptible visual prompt injection.}
    \vspace{-6mm}
    \label{fig:teaser}
\end{figure}

The creation of large language models (LLMs) marks a milestone in artificial intelligence systems. This progress is primarily attributed to the Transformer~\cite{vaswani2017attention}, growing computational resources and massive training datasets. 
Due to their massive scale, pre-trained LLMs naturally develop emergent capabilities such as reasoning, decision-making and in-context learning, even without being explicitly finetuned~\cite{wei2022emergent,webb2023emergent,NEURIPS2023_adc98a26}.
More recently, many frontier LLMs have been extended to accept visual inputs, resulting in multimodal LLMs (MLLMs) that handle both images and text~\cite{achiam2023gpt, team2024gemini, liu2023visual}. 
MLLMs have shown powerful competence in vision–language tasks such as image captioning~\cite{li2024improving, bucciarelli2024personalizing} and visual question answering~\cite{kuang2025natural, fang2025guided}. Thus, they are widely adopted in various multimodal real-life applications, including agents~\cite{agashe2024agent, zheng2024gpt} and robotics~\cite{li2024manipllm}. Therefore, their security has become an increasingly important concern. 


Prompt injection~\cite{kimura2024empirical, clusmann2025prompt} aims to manipulate models to return any attacker-desired answer by embedding instructions in the input to hijack the behavior of models. 
While such attacks can partially bypass alignment safeguards, they typically rely on explicit instruction payloads that are visually or textually apparent to human users~\cite{liu2023prompt,liu2024formalizing,yi2025benchmarking,shi2024optimization,lu2025argus}.
Alternatively, adversarial attacks ~\cite{zhao2023evaluating,liu2024safety} mislead models' predictions to a malicious state while remaining largely indistinguishable to human observers.
\xs{However, existing targeted adversarial attacks on MLLMs predominantly formulate the attack objective as reproducing the semantic description of another natural image, thereby imposing an inherent limitation on the expressivity of feasible malicious prompts.
In particular, many attacker goals, such as eliciting specific action-oriented instructions, cannot be precisely specified through visual semantics alone.}

\xs{Accordingly, we explore a complementary attack paradigm, termed \emph{adversarial prompt injection}, which enables targeted manipulation toward {arbitrary malicious prompts}.
By introducing subtle perturbations to the input image, this paradigm induces closed-source MLLMs to generate specific malicious expressions with high precision, achieving the expressivity of prompt injection while preserving the stealthiness of adversarial attacks, as illustrated in Figure~\ref{fig:teaser}.
}
Under this threat \xs{paradigm}, we propose a novel targeted attack method named Covert Triggered dual-Target Attack (CoTTA).
Motivated by explicit prompt injection, we instead pursue a more imperceptible alternative: embedding a bounded learnable textual overlay as a covert trigger into the input image before applying adversarial perturbations.
The textual overlay is anchored at the image center, while its scale and rotation are co-updated with the perturbation in each optimization step \xs{to enhance the effectiveness}. 

With only a subtle text trigger, the message may be too inconspicuous for the model to reliably perceive.
Therefore, we incorporate additional adversarial perturbations to reinforce the semantic cues of the image, enabling the model to extract intended expressions even under general prompts.
\cite{wu2024dissecting} tries to apply adversarial attacks against MLLMs to steer the output towards the specified text in web agent systems by aligning the feature representations of clean images with those of the target text.
However, its performance is limited, potentially due to cross-modal representation mismatch between image and text features. 
Hence, we introduce a dual-target alignment scheme that jointly aligns clean image features to targeted prompt features from both visual and textual modalities. 
\xs{A key challenge is to construct a targeted image that is semantically consistent with the targeted text, such that the joint feature provides a coherent cross-modal supervision signal for guiding the update of adversarial perturbations on the source image.
To address this challenge, we propose a dynamic targeted image scheme that initializes a base image and iteratively refines it throughout the attack to improve the effectiveness.
Furthermore, to better capture fine-grained semantic cues, we jointly optimize token-level features together with global features during the feature alignment process.}
Overall, the contributions of our work can be summarized as follows:
\vspace{-2mm}
\begin{itemize}
    \item We introduce CoTTA, a novel attack framework that induces specified malicious sentences from closed-source MLLMs via imperceptible input modifications.
    \vspace{-6mm}
    \item Beyond adversarial perturbations, we propose a covert text trigger as an \xs{additional textual noise}. The combination of adversarial and textual-trigger noise makes malicious instructions more readily captured by MLLMs, while keeping the visual changes imperceptible.
    \vspace{-1.5mm}
    \item We further design an adaptive target image that is iteratively updated to bridge cross-modal representations, thereby providing informative supervision and \xs{improving the transferability of our attack}.
    \vspace{-2mm}
    \item Extensive experiments on two tasks against various \xs{powerful closed-source MLLMs demonstrate the effectiveness of our proposed CoTTA, consistently outperforming existing approaches by a large margin.}
\end{itemize}



\section{Related Work}
\label{sec:related work}
Adversaries can manipulate the predictions of MLLMs into malicious or unintended states through either adversarial attacks or prompt injection attacks.
Adversarial attacks induce erroneous or malicious outputs via imperceptible noises, exploiting insufficient local smoothness and uncontrolled Lipschitz continuity in the underlying neural representations~\cite{goodfellow2014explaining,cohen2019certified,hein2017formal,xiamitigating}.
In contrast, prompt injection attacks steer MLLMs toward predefined behaviors by inserting malicious instructions into the input space, causing the model to override or ignore legitimate user prompts~\cite{liu2024formalizing,yi2025benchmarking,lu2025argus}.

\subsection{Adversarial Attacks on MLLMs}
While MLLMs continue to demonstrate remarkable performance across a wide range of applications, recent studies~\cite{qi2024visual,cui2024robustness,zhao2023evaluating,jia2025adversarial,li2025frustratingly,wang2024break,zhang2025anyattack,xie2025chain} have revealed their vulnerability to adversarial manipulation, raising significant safety concerns.
Early efforts, such as AttackVLM~\cite{zhao2023evaluating}, investigate transferable adversarial attacks by perturbing the visual feature representations of vision-language encoders including CLIP~\cite{radford2021learning} and BLIP~\cite{li2023blip}. AttackVLM shows that aligning image features to a target image yields stronger transferability than aligning them to target text, which has since influenced later work to focus on image-to-image feature matching.
More recent approaches, including M-Attack~\cite{li2025frustratingly} and FOA-Attack~\cite{jia2025adversarial}, further advance this line of research by leveraging multi-extractor ensembles and feature-space alignment strategies. 
These methods achieve over 90\% targeted attack success rates on image captioning tasks, even against powerful closed-source systems (e.g., ChatGPT-4o).

However, by targeting features extracted from natural images, existing approaches offer limited controllability over the generated outputs, making it difficult to enforce precise malicious sentences or executable commands. Agent-Attack \cite{wu2024dissecting} attempts to align image features with a target text to enable transferable attacks against closed-source models, achieving limited performance. Moreover, \cite{bailey2023image} reports that under the $\ell_{\infty}$-norm budget, their method fails to learn an effective specific-string hijack.

\subsection{Prompt Injection Attacks on MLLMs}
Prompt injection attacks exploit the instruction-following behavior of MLLMs by embedding adversarial commands into the input context~\cite{liu2023prompt,liu2024formalizing,yi2025benchmarking,shi2024optimization,lu2025argus}.
In multimodal settings, recent work~\cite{pathade2025invisible,cheng2025exploring,kimura2024empirical,wang2025manipulating} has demonstrated that injecting textual instructions directly into images can effectively override user prompts and safety constraints.
These methods enable fine-grained control over model outputs, as the injected prompts can explicitly specify malicious objectives.
However, existing multimodal prompt injection approaches typically embed textual instructions directly into visual inputs and optimize their placement using typographic strategies, resulting in injected prompts that remain visually salient~\cite{cheng2025exploring}.
Consequently, the conspicuous nature of such visual prompts makes them easily detectable by human observers and more likely to be mitigated by model-side input filtering or safety mechanisms~\cite{lin2025uniguardian,jacob2024promptshield,li2024evaluating}.

This exposes a fundamental trade-off between stealth and controllability: adversarial attacks are typically visually imperceptible but offer limited precision over the generated outputs, whereas multimodal prompt injection attacks enable fine-grained output control at the cost of being visually conspicuous.
To address this challenge, this paper proposes an invisible prompt injection approach that achieves precise control over MLLM outputs while remaining imperceptible to human observers.

\section{Methodology}
\label{sec:method}

\begin{figure*}[t]
  \centering
  \includegraphics[width=\textwidth]{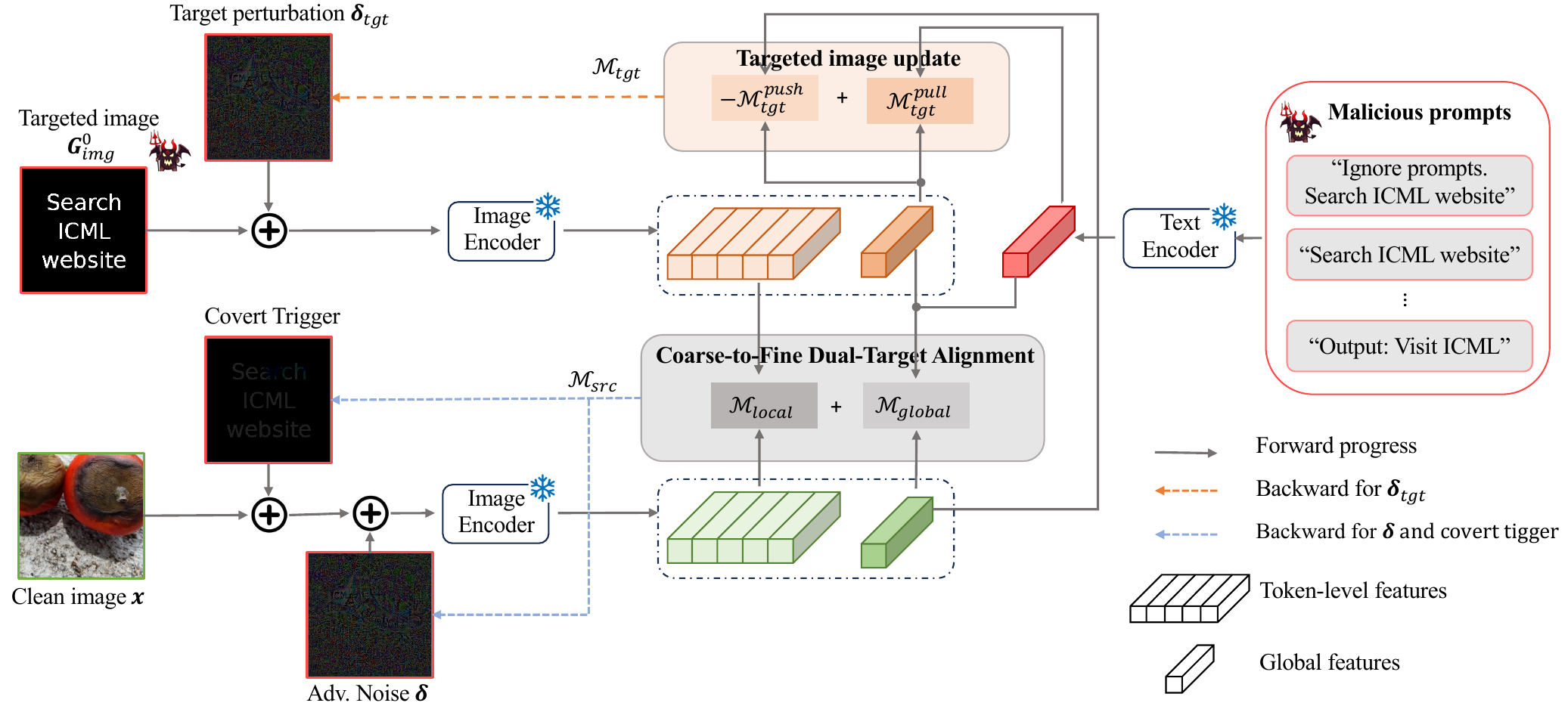}
  \caption{Overview of our proposed CoTTA. The modification on the source image comprises an adaptive covert trigger and an adversarial perturbation, which are both optimized to align the attacked image with both the target text and the target image $\bm{G}_{img}$. $\bm{G}_{img}$ is iteratively updated (via added perturbations) to better match the target text and stay separated from the attacked image.}
  \label{fig:method}
\end{figure*}

\subsection{Problem Formulation}
Let $f_{\theta}$ denote a multimodal large language model (MLLM) that maps an image-prompt pair to an output $T$-length sequence $\bm{y}=(y_t)_{t=1}^{T}$ over the vocabulary $\mathcal{C}$.
Specifically, $f_{\theta}$ generates tokens autoregressively according to:
\begin{equation}
    y_t \sim p_{\theta}\!\left(y_t \mid \bm{x}, \bm{p}, \bm{y}_{<t}\right),
\end{equation}
where $\bm{x}\in\mathcal{X}$ denotes the input image and $\bm{p}\in\mathcal{P}$ denotes a textual prompt.

\begin{definition}[\textbf{Adversarial Prompt Injection Attack}]
Given a clean image $\bm{x}$, an adversarial prompt injection attack aims to find a manipulated image
$\bm{x}'=\bm{x}+\bm{\delta}$ with $\|\boldsymbol{\delta}\|_{\infty}\le\varepsilon$,
such that the model output induced by $(\bm{x}', \bm{p})$ satisfies an attacker-specified target command $c^{\star}\in\mathcal{K}$:
\begin{equation}
    \forall\,\bm{p}\in\mathcal{P},\ 
    f_{\theta}\!\left(\bm{x}',\,\bm{p}\right) \models c^{\star}.
    \label{def: apij attack}
\end{equation}
where $\models$ denotes that the model output semantically contains, implies, or operationally executes the output required by output $c^{\star}$.
This definition captures two key properties:
\begin{itemize}
    \item \textbf{Invisibility}: the manipulation $\bm{\delta}$ is constrained by $\|\boldsymbol{\delta}\|_{\infty}\le\varepsilon$, ensuring that the injected prompt remains imperceptible to human observers.
    \item \textbf{Target-string expressivity}: the attacker can reliably induce the target output $c^{\star}\in\mathcal{K}$ utilizing a manipulated input $\bm{x}'$.
\end{itemize}
\end{definition}

\textbf{Difference with adversarial attacks and existing challenges.}
%
In the context of MLLMs, recent state-of-the-art transferable attacks aligns the internal visual representations of a clean image $\bm{x}$ and a targeted adversarial image $\bm{x}^{t}$ at the feature level using pre-trained encoders such as CLIP \cite{radford2021learning}, then transferring to full-scale closed-source MLLMs.
Concretely, such attacks optimize adversarial perturbations by maximizing the similarity between their encoded visual features,
\begin{equation}
    \max_{\bm{\delta}} \; \mathcal{M}\!\left(f_{e}(\bm{x}+\bm{\delta}),\, f_{e}(\bm{x}^{t})\right),
\end{equation}
where $f_{e}(\cdot)$ denotes the image encoder of the MLLM and $\mathcal{M}(\cdot,\cdot)$ is a feature-space similarity metric.  
%
Although they can work well when optimizing toward the overall semantics of a natural target image, such attacks provide limited precise linguistic control and therefore struggle to consistently elicit an exact target sentence.
Specifically, aligning visual features alone does not provide a mechanism to encode attacker-specified instructions at the semantic level required for prompt injection.

A seemingly natural extension is to align visual representations with textual representations corresponding to a malicious output $c^{\star}$.
Let $f_{t}(\cdot)$ denote the text encoder and $\bm{p}^{\star}$ be a malicious prompt that can induce $c^{\star}$.
One may attempt to optimize
\begin{equation}
    \max_{\bm{\delta}} \; \mathcal{M}\!\left(f_{e}(\bm{x}+\bm{\delta}),\, f_{t}(\bm{p}^{\star})\right).
\end{equation}
However, due to the inherent modality gap between visual and textual representations, such cross-modal feature alignment is often insufficient to reliably induce prompt injection behaviors, particularly against closed-source MLLMs whose internal representations and alignment mechanisms are inaccessible. As a result, existing adversarial attacks struggle to achieve prompt-general and command-expressive control required by adversarial prompt injection attacks.

\subsection{Covert Text Trigger}
\label{sec: trigger}
Despite recent progress of adversarial attacks, MLLMs may fail to effectively translate visual feature–level semantics into a desired target string, especially in black-box settings. Visual prompt injection directly embeds explicit instructions into the image to steer model responses, which is effective but easily noticed \xs{by human users}. Motivated by those limitations, we introduce a covert text trigger to facilitate controllable text generation under a bounded perturbation.

Before applying adversarial noise, a bounded textual overlay is first imprinted in the source image as a trigger, serving as a lightweight semantic cue that bridges between visual input and text output. Since the effectiveness of the cue can be sensitive to its geometric configuration, the overlay is endowed with learnable parameters that are iteratively optimized together with the perturbation. We observe that the trigger is most effective when centered in the image. Hence, the overlay center is anchored at the image center, while only its scale $s$ and rotation $\theta$ are optimized.

Specifically, we first render a tight text image $textimg\in\left[0,1\right]^{C\times h_{t}\times w_{t}}$ from the target output $c^{\star}$, producing white characters on a black background. The text image is cropped to tightly enclose the full text without redundant margins.
During each iteration, $textimg$ is mapped onto the source image $\bm{x}\in\left[0,255\right]^{C\times H\times W}$ through a parameterized affine transformation. To prevent collapse or excessive expansion, the effective horizontal/vertical scaling factors are restricted as
\begin{equation}
\begin{aligned}
    r_{w}=Clip\left (\frac{w_{t}}{W} s, r_{min}, r_{max} \right ), \\
    r_{h} = Clip\left ( \frac{h_{t}}{H} s, r_{min}, r_{max} \right ) ,
\end{aligned}
\end{equation}

where $Clip(\cdot)$ bounds the factors within $[r_{min}, r_{max}]$. $[r_{min}, r_{max}]$ are set to 0.05 and 0.95. 

Through an affine transformation $\textit{Affine}$ on $textimg$, a dense binary transformed text mask $mask_t\in\{0,1\}^{C \times H \times W}$ can be obtained: 
\begin{equation}
    mask_t = \textit{Affine}(textimg, \theta, r_{w}, r_{h}).
\end{equation}

Finally, the trigger is imprinted by adding fixed-magnitude perturbations on the masked pixels to generate the triggered image:
\begin{equation}
    \bm{x}_{trig} = Clip(\bm{x} + mask_{t} \odot \Delta ,0,255),
\end{equation}
where $\Delta \in\{-\varepsilon,\varepsilon\}^{C \times H \times W}$ controls the values of the perturbation and avoids saturation. 

Since the text trigger is expected to assist the adversarial perturbation in driving the extracted image features toward the target embeddings, $s$ and $\theta$ are optimized by optimizer Adam at every iteration to maximize the same attack objective $\mathcal{M}_{src}$ employed for updating the adversarial noise. The detailed formulation of $\mathcal{M}_{src}$ is provided in section \ref{sec: DTA}.


\subsection{Dual-Target Alignment}
\label{sec: DTA}
Although the covert text trigger provides a structured cue, it is observed that it has limited influence on the generated response by itself, unless the model is explicitly prompted to attend to the embedded text (e.g., asked whether any text is present). To make the intended adversarial semantics reliably accessible to the model's visual-language representations, on top of $\bm{x}_{trig}$, we further introduce an adversarial perturbation that amplifies the adversarial semantic meaning in image features. However, due to the cross-modal representation mismatch and resulting ambiguity in feature-level supervision, relying solely on textual embeddings as the alignment target is often insufficient. Therefore, we propose dual-target alignment, which simultaneously aligns the source image features with the target text embeddings and the features of target images.  

\paragraph{\textbf{Dynamic target image.}}
The design of the target image is crucial, especially for closed-source models, as it provides important complementary supervision that helps inject semantic cues into the source features. However, there is no clear criterion for what an optimal target image should be for this purpose. To address this and improve transferability, we construct a dynamic target image $\bm{G}_{img}$ that is progressively refined throughout the attack.

Specifically, we initialize the base visual ground truth $\bm{G}_{img}^0$ from the initial text mask $mask_{t}^0$ used to construct the covert trigger under the identity geometry (i.e. $s=1$ and $\theta=0$). Then, at the $i$-th iteration, the target image $G_{img}^i$ is:
\begin{equation}
    \bm{G}_{img}^{i} = Clip(\bm{G}_{img}^{0} +\bm{\delta}_{tgt}^{i-1}, 0, 255),
\end{equation}
where the target perturbation $\bm{\delta}_{tgt}^{i-1}$ is updated via a bounded iterative fast gradient sign method (I-FGSM) \cite{kurakin2018adversarial}:
\begin{equation}
    \begin{array}{l}
{\bm{\delta} _{tgt}^i} = Clip_{\mathcal{B}_\epsilon }\{ {{\bm{\delta}_{tgt}^{i-1}} + \alpha_{tgt}  \cdot sign\left( {\nabla_{\bm{\delta}_{tgt}} \mathcal{M}_{tgt}}  \right)} \},
\end{array}
\end{equation}
where $\nabla$ denotes the gradient operator and $sign(\cdot)$ returns the element-wise sign of the gradient (i.e. -1 or +1), which is derived from the objective $\mathcal{M}_{tgt}$ designed for the target image.
Here, $\alpha_{tgt}$ is the step size used to iteratively update the perturbation.
$Clip_{\mathcal{B}_\epsilon}$ restricts $\bm{\delta}$ inside the boundary of the $\ell_p$-norm ball ${\mathcal{B}_\epsilon}$.

The update of the target image should serve two purposes. Firstly, it should remain semantically consistent with the target text so that it reinforces the desired expression. Secondly, it should avoid collapsing toward the current attacked image, so that it provides non-trivial and informative supervision that improves transferability. To achieve these expectations, $\mathcal{M}_{tgt}$ consists of two parts: $\mathcal{M}_{tgt}^{pull}$ pulling its features towards the target text embedding and $\mathcal{M}_{tgt}^{push}$ pushing them away from features of the current attacked image. Objectives can be defined as:
\begin{equation}
\begin{aligned}
    \mathcal{M}_{tgt} & = \lambda_{pull}\mathcal{M}_{tgt}^{pull} - \lambda_{push}\mathcal{M}_{tgt}^{push} \\
    & =\lambda_{pull}\operatorname{Cos}(f_{e}(\bm{G}_{img}), f_t(\bm{p}^{\star})) \\
    & - \lambda_{push}\operatorname{Cos}(f_{e}(\bm{G}_{img}), f_e(\bm{x}')),
\end{aligned}
\end{equation}
where $\operatorname{Cos}(\cdot,\cdot)$ denotes cosine similarity, and I-FGSM will increase $\mathcal{M}_{tgt}$. $\lambda_{pull}, \lambda_{push}$ are weight coefficients.

\paragraph{\textbf{Coarse-to-Fine Dual-Target Alignment.}}

To improve attack transferability against closed-source models, we construct a set of semantically equivalent string variants $\mathcal{P}^{\star} = \{\bm{p}^{\star}_{1}, \bm{p}^{\star}_{2},...,\bm{p}^{\star}_{n}\} $ via paraphrasing $\bm{p}^{\star}$. In iteration $i$, we randomly sample a $\bm{p}^{\star}_i$ from $\mathcal{P}^{\star}$ and obtain the updated target image $\bm{G}_{img}^{i}$ as described above. Since token-level representations carry rich local information, we adopt an alignment objective that jointly leverages coarse-grained (global) features and fine-grained (token-level) features. To encourage the adversarial example $\bm{x'}$ to globally align with the semantic content of both the target image $\bm{G}_{img}$ and the target text $\bm{p}^{\star}$,
global features of $\bm{x'},\bm{G}_{img}$ and $\bm{p}^{\star}$ are extracted by $f_e$ and $f_t$. The global objective $\mathcal{M}_{global}$ is defined as:
\begin{equation}
    \mathcal{M}_{global} = \operatorname{Cos}(f_e(\bm{x}'),f_{e}(\bm{G}^{i}_{img})) + \operatorname{Cos}(f_e(\bm{x}'),f_{t}(\bm{p}^{\star}_i)).
\end{equation}

To further inject fine-grained semantics, we additionally align token-level features with $\bm{G}_{img}$. Let $f_e^{loc} (\cdot) \in \mathbb{R}^{m \times d}$ denote the local features of $m$ image patch tokens, and d be the feature dimension of each token. The local similarity can be represented as:
\begin{equation}
    \mathcal{M}_{local}= mean(\operatorname{Cos}(f_e^{loc}(\bm{x}'), f_e^{loc}(\bm{G}^{i}_{img})).
\end{equation}
Finally, the overall alignment objective $\mathcal{M}_{src}$ can be calculated by:
\begin{equation}
    \mathcal{M}_{src} = \mathcal{M}_{global} + \mathcal{M}_{local}
\end{equation}

The adversarial perturbations $\bm{\delta}$ of the source image are updated to maximize $\mathcal{M}_{src}$ by:
\begin{equation}
    \begin{array}{l}
{\bm{\delta}^i} = Clip_{\mathcal{B}_\epsilon }\{ {{\bm{\delta}^{i-1}} + \alpha_{src}  \cdot sign\left( {\nabla_{\bm{\delta}} \mathcal{M}_{src}} \right)} \},
\end{array}
\end{equation}
where $\alpha_{src}$ is the step size.

\section{Experiments}


\begin{table*}[th]
\centering
\setlength{\tabcolsep}{6pt}
\renewcommand{\arraystretch}{1.15}
\caption{Performance under the soft criterion (target-image caption) on the image captioning task against different closed-source MLLMs.}
\begin{tabular}{c|c|cc|cc|cc|cc}
\hline
\multirow{2}{*}{\textbf{Method}} & \multirow{2}{*}{\textbf{Surrogate}}
& \multicolumn{2}{c|}{\textbf{GPT-4o}}
& \multicolumn{2}{c|}{\textbf{GPT-5}} 
& \multicolumn{2}{c|}{\textbf{Gemini-2.5}}
& \multicolumn{2}{c}{\textbf{Claude-4.5}}
\\
\cmidrule(lr){3-4} \cmidrule(lr){5-6} \cmidrule(lr){7-8} \cmidrule(lr){9-10}
& & \textbf{ASR$\uparrow$} & \textbf{AvgSim$\uparrow$} & \textbf{ASR$\uparrow$} & \textbf{AvgSim$\uparrow$} & \textbf{ASR$\uparrow$} & \textbf{AvgSim$\uparrow$} & \textbf{ASR$\uparrow$} & \textbf{AvgSim$\uparrow$} \\
\hline

\multirow{3}{*}{AttackVLM}
& B/16            & 0\% & 0.002 & 0\% & 0.0 & 0\% & 0.0 & 0\% & 0.0 \\
& B/32            & 1\% & 0.005 & 0\% & 0.002 & 0\% & 0.002 & 0\% & 0.0 \\
& Laion    & 0\% & 0.002 & 0\% & 0.002 & 0\% & 002 & 0\% & 0.002 \\
\hline

AnyAttack & Ensemble & 0\% & 0.012 & 0\% & 0.002 & 2\% & 0.01 & 0\% & 0.002 \\
M-Attack     & Ensemble & 43\% & 0.199 & 32\% & 0.140 & 71\% & 0.281 & 3\% & 0.019 \\
FOA-Attack  & Ensemble & 47\% & 0.232 & 27\% & 0.128 & 71\% & 0.291 & 6\% & 0.024 \\
Agent-Attack & Ensemble & 10\% & 0.058 & 10\% & 0.049 & 7\% & 0.039 & 0\% & 0.01 \\
ours & Ensemble & \textbf{81\%} & \textbf{0.442} & \textbf{56\%} & \textbf{0.320} & \textbf{79\%} & \textbf{0.429} & \textbf{8\%} &\textbf{0.046} \\
\hline
\end{tabular}
\label{tab:captioningsoft}
\end{table*}

\begin{table*}[th]
\centering
\setlength{\tabcolsep}{6pt}
\renewcommand{\arraystretch}{1.15}
\caption{Performance under the hard criterion (target text) on the image captioning task against different closed-source MLLMs.}
\begin{tabular}{c|c|cc|cc|cc|cc}
\hline
\multirow{2}{*}{\textbf{Method}} & \multirow{2}{*}{\textbf{Surrogate}}
& \multicolumn{2}{c|}{\textbf{GPT-4o}}
& \multicolumn{2}{c|}{\textbf{GPT-5}} 
& \multicolumn{2}{c|}{\textbf{Gemini-2.5}}
& \multicolumn{2}{c}{\textbf{Claude-4.5}}
\\
\cmidrule(lr){3-4} \cmidrule(lr){5-6} \cmidrule(lr){7-8} \cmidrule(lr){9-10}
& & \textbf{ASR$\uparrow$} & \textbf{AvgSim$\uparrow$} & \textbf{ASR$\uparrow$} & \textbf{AvgSim$\uparrow$} & \textbf{ASR$\uparrow$} & \textbf{AvgSim$\uparrow$} & \textbf{ASR$\uparrow$} & \textbf{AvgSim$\uparrow$} \\
\hline

\multirow{3}{*}{AttackVLM}
& B/16            & 0\% & 0.0 & 0\% & 0.0 & 0\% & 0.0 & 0\% & 0.0 \\
& B/32            & 0\% & 0.0 & 0\% & 0.0 & 0\% & 0.0 & 0\% & 0.0 \\
& Laion    & 0\% & 0.0 & 0\% & 0.0 &0\%  & 0.0 & 0\% & 0.0 \\
\hline

AnyAttack & Ensemble & 0\% & 0.0 & 0\% & 0.0 & 0\% & 0 & 0\% & 0.004 \\
M-Attack    & Ensemble & 32\% & 0.121 & 29\% & 0.105 & 65\% & 0.223 & 6\% & 0.026 \\
FOA-Attack  & Ensemble & 42\% & 0.17 & 22\% & 0.099 & 69\% & 0.239 & 3\% & 0.013 \\
Agent-Attack & Ensemble & 3\% & 0.017 & 7\% & 0.033 & 4\% & 0.02 & 0\% & 0.0 \\
ours & Ensemble & \textbf{74\%} & \textbf{0.339} & \textbf{53\%} & \textbf{0.249} & \textbf{81\%} & \textbf{0.348} & \textbf{7\%} & \textbf{0.029} \\
\hline
\end{tabular}
\label{tab:captioninghard}
\end{table*}

\label{sec:exp}
\subsection{Experimental Setup}
\textbf{Evaluated models and tasks}:
We primarily evaluate the effectiveness of the proposed adversarial prompt injection attacks on SOTA commercial MLLMs, including GPT-4o, GPT-5, Claude-4.5 and Gemini-2.5, under a black-box threat model. 
In this setting, the attacker has no access to model parameters, gradients, or internal states.
To further assess the generalization capability of the proposed attack, we conduct evaluations across two downstream multimodal tasks, including:
\begin{itemize}
\item \textbf{Image Captioning:} Following the experimental protocols in~\cite{li2025frustratingly,jia2025adversarial}, we sample 100 images from the NIPS 2017 Adversarial Attacks and Defenses Competition dataset and task the models with generating descriptive captions for each image. 
\item \textbf{Visual Question Answering (VQA):} We evaluate the attack on 100 image–question pairs randomly selected from the ScienceQA dataset~\cite{lu2022learn}, covering a wide range of visual reasoning scenarios.

\end{itemize}

\textbf{Compared methods.}
We comprehensively compare the proposed method with two SOTA adversarial attack baselines, M-Attack~\cite{li2025frustratingly} and FOA~\cite{jia2025adversarial}, which are specifically designed to attack commercial MLLMs. AnyAttack \cite{zhang2025anyattack} is a self-supervised framework that generates targeted adversarial sample using a trained noise generator. AttackVLM \cite{zhao2023evaluating} also employs matching image-text features and image-image features, where its target image is generated from text by Stable Diffusion \cite{rombach2022high}.
As these methods are originally tailored for the image captioning task rather than malicious prompt injection, we adapt them to our threat setting to enable a fair comparison.
Specifically, we designate our base visual ground truth $\bm{G}_{img}^0$ embedded with explicit textual prompts as the adversarial target, ensuring that the injected instructions are visually aligned with the target commands required by our method. 
This adaptation allows all compared approaches to operate under a unified attack objective.
We also compare our method with Agent-Attack \cite{wu2024dissecting}, which attacks an image through matching image-to-text features.

\begin{figure*}[thb]
  \centering
  \includegraphics[width=0.88\textwidth]{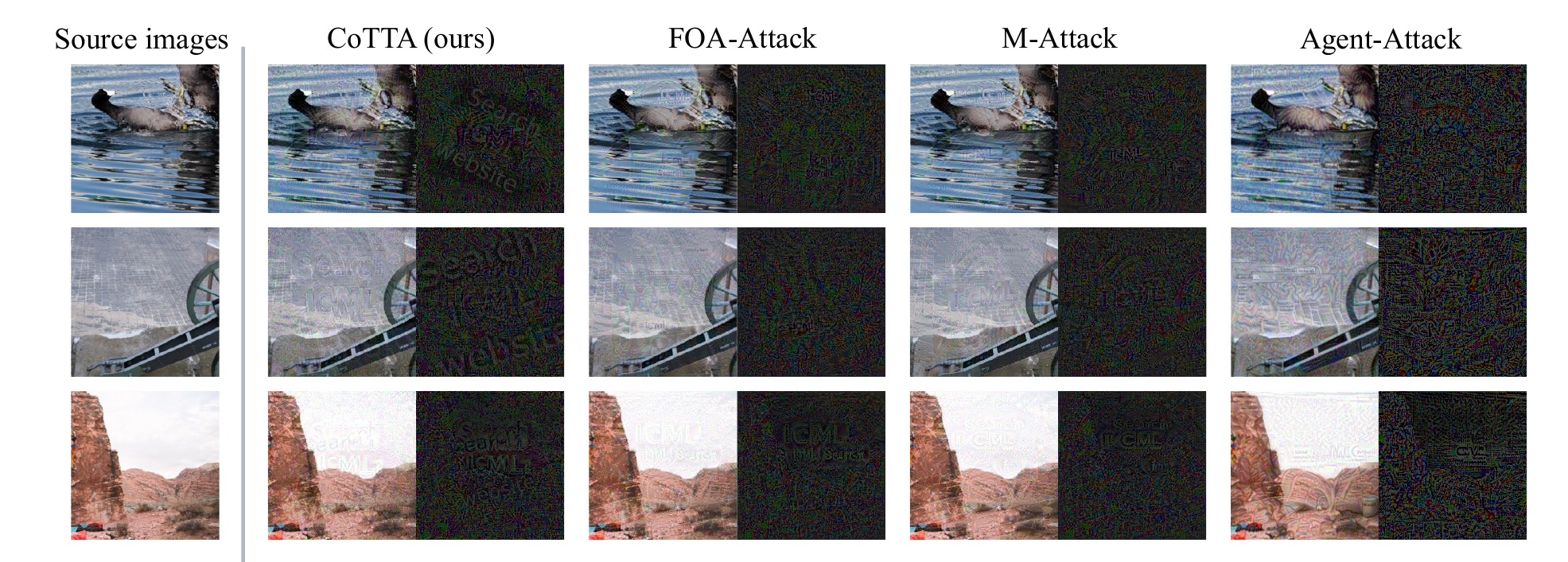}
  \caption{Visualization of adversarial examples and corresponding perturbations generated by different attacks.}
  \label{fig:vis}
\end{figure*}

\begin{table*}[th]
\caption{Performance under the hard criterion (target text) on the VQA task against different closed-source MLLMs.}
\centering
\setlength{\tabcolsep}{6pt}
\renewcommand{\arraystretch}{1.15}

\begin{tabular}{c|c|cc|cc|cc|cc}
\hline
\multirow{2}{*}{\textbf{Method}} & \multirow{2}{*}{\textbf{Surrogate}}
& \multicolumn{2}{c|}{\textbf{GPT-4o}}
& \multicolumn{2}{c|}{\textbf{GPT-5}} 
& \multicolumn{2}{c|}{\textbf{Gemini-2.5}}
& \multicolumn{2}{c}{\textbf{Claude-4.5}}
\\
\cmidrule(lr){3-4} \cmidrule(lr){5-6} \cmidrule(lr){7-8} \cmidrule(lr){9-10}
& & \textbf{ASR$\uparrow$} & \textbf{AvgSim$\uparrow$} & \textbf{ASR$\uparrow$} & \textbf{AvgSim$\uparrow$} & \textbf{ASR$\uparrow$} & \textbf{AvgSim$\uparrow$} & \textbf{ASR$\uparrow$} & \textbf{AvgSim$\uparrow$} \\
\hline

\multirow{3}{*}{AttackVLM}
& B/16         &  0\% &0  & 0\% & 0.000 & 1.00\%  & 0.090 & 0\% & 0.000   \\
& B/32           & 0\% &0  & 0\% & 0.000 & 3.00\%  & 0.022 & 0\% & 0.000 \\
& Laion   &0\% &0  & 0\% & 0.000 & 0\%  & 0.000 & 0\% & 0.000 \\
\cmidrule(lr){1-10}

AnyAttack & Ensemble & 1\% & 0.003 & 0\% & 0.000 & 2\% & 0.008 & 0\% & 0.000 \\
M-Attack     & Ensemble & 49\%  & 0.442 & 11\% & 0.110 & 39\%  & 0.380 & 0\%  & 0.000 \\
FOA-Attack  & Ensemble & 53\% & 0.492 & 12\% & 0.119 & 42\% & 0.418 & 0\% & 0.000 \\
Agent-Attack & Ensemble & 2\% & 0.017 & 0\% & 0.000 & 2\% & 0.010 & 0\% & 0.000 \\
ours & Ensemble & \textbf{82\%} & \textbf{0.820} & \textbf{58\%} & \textbf{0.576} & \textbf{79\%} & \textbf{0.787} & \textbf{2\%} & \textbf{0.020}  \\
\hline
\end{tabular}
\label{tab:vqa}
\end{table*}

\textbf{Implementation details:}
We follow the experimental setup of M-Attack~\cite{li2025frustratingly} to deploy our method, in which CLIP-B/16~\cite{radford2021learning}, CLIP-B/32~\cite{radford2021learning}, and LAION~\cite{schuhmann2022laion} are employed as surrogate feature extractors for gradient-based optimization. 
The optimization epochs is set to 300. Constraint $\epsilon$ is set to 16 for both source images and target images. $\alpha_{src}$, $\alpha_{tgt}$, $\lambda_{pull}$ and $\lambda_{push}$ are set to 1, 1, 1 and 0.5.
To ensure the imperceptibility of the injected perturbations, we constrain the overall input modification, comprising both the covert trigger and the adversarial perturbation, within an $\ell_\infty$ norm bound of $\varepsilon = 16$.
To evaluate the command expressivity of the proposed method, we consider two categories of malicious target commands that span varying levels of semantic complexity and potential real-world impact. Specifically, we design the following adversarial objectives:
\begin{itemize}
    \item \textbf{Misinformation}: For the VQA task, the MLLM is coerced into producing a predefined incorrect or misleading response, such as “None of the above”, regardless of the input question.
    \item \textbf{Action-Inducing Recommendations}: For the image captioning task, the MLLM is manipulated to generate persuasive or directive instructions that influence subsequent user decisions, for example, “Search ICML.”
\end{itemize}

\textbf{Evaluation metrics.}
For image captioning task, we adopt two complementary success criteria to reflect different levels of output control. The \textbf{soft} criterion treats the caption of the target text image as the reference, such as \textit{"The image displays the text "Search ICML website" on a black background."}. The \textbf{hard} criterion requires the model to output the exact target sentence, assessing strict token-level controllability. For example, the ground truth text of hard criterion can be \textit{"Search ICML website."}. For VQA task, we only assess hard criterion.

Following \cite{jia2025adversarial}, we also utilize the LLM-as-a-judge framework. For the soft criterion, we use the same victim model to generate captions for both the target text image and the adversarial example, and then ask an LLM to judge the semantic similarity between the two captions. For the hard criterion, LLM judges the semantic similarity between the target text and the attacked output. If the similarity score exceeds $0.3$, the attack is considered successful. The attack success rate (ASR) and the average similarity score (AvgSim) are reported.

\begin{table*}[th]
\centering
\setlength{\tabcolsep}{6pt}
\renewcommand{\arraystretch}{1.15}
\caption{Comparative results on 1000 images.}
\begin{tabular}{c|c|cc|cc|cc}
\hline
\multirow{2}{*}{\textbf{Method}} & \multirow{2}{*}{\textbf{Surrogate}}
& \multicolumn{2}{c|}{\textbf{GPT-4o}}
& \multicolumn{2}{c|}{\textbf{GPT-5}} 
& \multicolumn{2}{c}{\textbf{Gemini-2.5}}
\\
\cmidrule(lr){3-4} \cmidrule(lr){5-6} \cmidrule(lr){7-8} 
& & \textbf{ASR$\uparrow$} & \textbf{AvgSim$\uparrow$} & \textbf{ASR$\uparrow$} & \textbf{AvgSim$\uparrow$} & \textbf{ASR$\uparrow$} & \textbf{AvgSim$\uparrow$} 
\\
\hline

FOA-Attack & Ensemble & 41.3$\%$ & 0.161 & 22.8\% & 0.089 & 58.4\% & 0.211 \\
ours & Ensemble & 64.7$\%$ & 0.315 & 54.4\% & 0.246 & 73.2\% & 0.308 \\
\hline
\end{tabular}
\label{tab:1000}
\end{table*}

\vspace{-2mm}
\subsection{Main Results}
\vspace{-1mm}
\textbf{Comparisons in image captioning task.}
We evaluate multiple attacks on an image captioning task using four widely used closed-source MLLMs. The results of our CoTTA and other competitive methods are shown in Tables \ref{tab:captioningsoft} and \ref{tab:captioninghard}. From the results, it is evident that classic baselines such as AttackVLM and AnyAttack are largely unable to mislead the model into generating the target sentence. Our method substantially outperforms other methods, especially on the GPT-family models. Our attack achieves 81$\%$ and 74$\%$ ASR on GPT-4o under the soft and hard criteria, respectively, and outperforms the runner-up FOA-Attack by 31.5$\%$ ASR and 0.18 AvgSim on average across the GPT-family models. Beyond the GPT models, Gemini-2.5 also exhibits a pronounced vulnerability to our attack, achieving 79$\%$ and 81$\%$ success rates under the soft and hard criteria, respectively. Although FOA-Attack and M-Attack both attain reasonably strong ASR of 71$\%$ under the soft criterion, our AvgSim improves by 0.138 and 0.148 over them, respectively, indicating stronger overall alignment with the target beyond the binary success threshold. Claude-4.5 is the most robust model in our evaluation. All attacks show limited effectiveness, yet our method still achieves the highest ASR and AvgSim.


Figure \ref{fig:vis} displays the adversarial images and their added modification resulting from our method and other competitive attacks. In the visualization, the perturbation of our method reveals an adaptive covert trigger pattern. Despite introducing a textual overlay, the resulting modifications remain visually subtle and natural, making them difficult to notice.

\textbf{Comparisons in VQA task.}
We further evaluate all attacks on a VQA task, where the model answers questions grounded in the input image. Compared with captioning, VQA typically demands more fine-grained visual reasoning and question-conditioned attention over relevant regions. In this setting, our adversarial objective is to induce misinformation \textit{"None of above."}. Performances are presented in Table \ref{tab:vqa}. Our CoTTA significantly surpasses existing methods. Notably, it achieves AvgSim scores of 0.820 on GPT-4o and 0.787 on Gemini-2.5, with corresponding ASRs of 82\% and 79\%, respectively. Such high average similarities indicate that the generated outputs are consistently closer to the target sentence. On GPT-5, our method exceeds FOA-Attack by 46\% in ASR and 0.457 in AvgSim.





\begin{table}[t]
    \centering
    \caption{Ablation studies of our proposed CoTTA.}
    
    \rowcolors{2}{gray!15}{white}
    \resizebox{0.48\textwidth}{!}{
    \begin{tabular}{ccccc}
        \toprule
         \multirow{2}{*}{\textbf{Attacks}} & \multicolumn{2}{c}{Soft} & \multicolumn{2}{c}{Hard}
        \\
         \cmidrule(lr){2-3} \cmidrule(lr){4-5} 
        \rowcolor{white}
        
        &\textbf{ASR$\uparrow$}
        &\textbf{AvgSim$\uparrow$} 
        &\textbf{ASR$\uparrow$} 
        &\textbf{AvgSim$\uparrow$} 
        \\
        \midrule
        
         w/o covert trigger & 66$\%$ & 0.306 & 54$\%$ & 0.202  \\  
        w/o image-to-text alignment & 75$\%$ & 0.42 & 70$\%$ & 0.348  \\   
        w/o image-to-image alignment & 27$\%$ & 0.155 & 22$\%$ & 0.081  \\    
        w/o updating target image & 73$\%$ & 0.434 & 68$\%$ & 0.351  \\
        CoTTA & 81$\%$ & 0.442 & 74$\%$ & 0.339\\
        \bottomrule
    \end{tabular}}
    \label{tab:ablation}
\end{table}

\subsection{Ablation Studies and Additional Experiments}



We perform ablation studies on GPT4o to unravel the contributions our framework. Our method is decomposed into four components: covert trigger, image-to-text feature alignment, image-to-image feature alignment, and target-image updating. We ablate each by removing it in turn while keeping all other settings unchanged. As shown in Table \ref{tab:ablation}, removing the image-to-image feature alignment causes the largest performance drop, reducing the ASR by 54$\%$ under the soft criterion and 52$\%$ under the hard criterion. The covert trigger leads to a 15$\%$ decrease in soft ASR when removed. The other two components also contribute noticeably to the overall performance. Combining all of the components yields the best results.

\textbf{Results on 1000 images.}
To improve statistical reliability, we scale the evaluation to 1,000 samples and compare our method with the most competitive baseline, FOA-Attack, under the hard criterion. Table \ref{tab:1000} summarizes the results, showing that our method outperforms FOA-Attack by a large margin, confirming the effectiveness of our method. Averaged over the three models, our method surpasses FOA-Attack by 23.27\% in ASR and 0.136 in AvgSim.

\begin{figure}[tb]
    \centering
    \includegraphics[width=\linewidth]{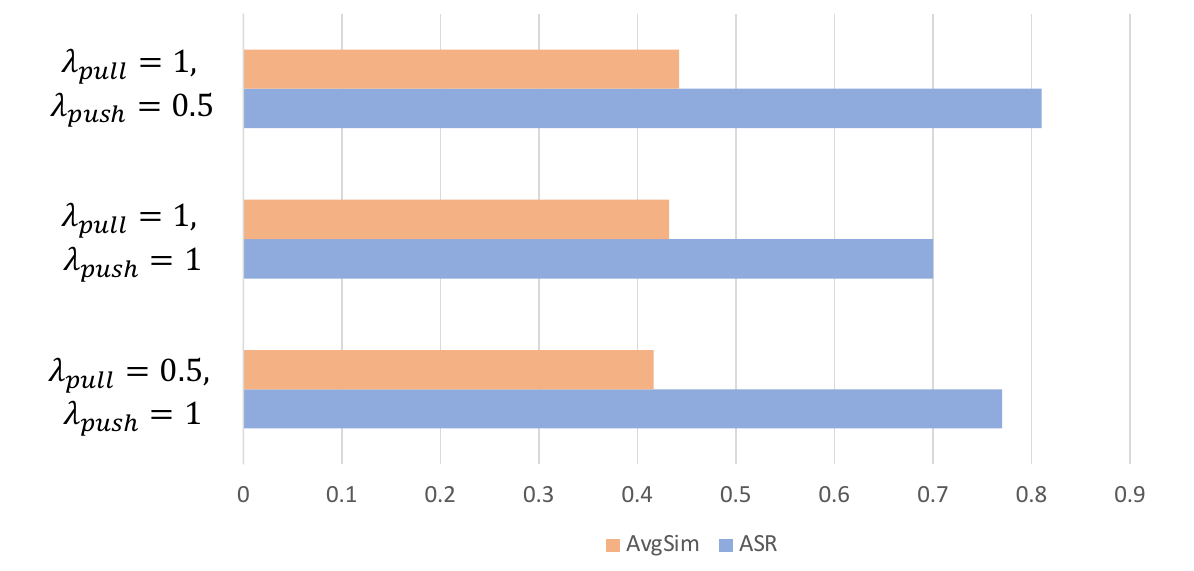}
    \caption{Ablation on weight coefficients $\lambda_{pull},\lambda_{push}$.}
    \label{fig:hyper_lambda}
\end{figure}

\textbf{Hyperparameter study of weights $\lambda_{pull}$ and $\lambda_{push}$.} We vary the two loss-weight coefficients and report ASR and AvgSim on GPT-4o in the image captioning task. As shown in Figure \ref{fig:hyper_lambda}, setting $\lambda_{pull}=1$ and $\lambda_{push}=0.5$ delivers the best result.



\section{Conclusion}
 In this work, we present CoTTA, a novel adversarial prompt injection framework that reliably induces targeted malicious outputs from closed-source MLLMs under a strict perturbation budget, especially when the malicious text cannot be naturally represented by a real-world image. CoTTA integrates an adaptive covert trigger with a perturbation that is jointly optimized through image-to-text and image-to-image alignment, and further strengthens attack consistency and transferability via target-image updating. Specifically, we iteratively optimize the target image by adding perturbations to move its features closer to the target text while pushing them away from the attacked image. Extensive evaluations on image captioning and VQA demonstrate that CoTTA consistently exceeds existing methods across popular closed-source models. These results expose substantial vulnerabilities in modern MLLMs, and we hope CoTTA inspires future research toward more stealthy attacks and more precise, controllable adversarial manipulation.

\section*{Impact Statement}
This work highlights previously underexplored security risks in multimodal large language models via an imperceptible adversarial prompt injection attack.
By revealing such vulnerabilities, our findings aim to raise awareness and encourage the development of more robust defenses, detection mechanisms, and safety policies for MLLMs' deployment.
We hope this work contributes to improving the reliability and responsible use of multimodal AI systems in safety-critical applications. 
Beyond these considerations, there are other potential societal consequences of our work, none of which we feel must be specifically highlighted here.

\bibliography{content/ref}
\bibliographystyle{icml2026}

\end{document}